\documentclass[11pt]{article}

\usepackage[preprint]{acl}

\usepackage{times}
\usepackage{latexsym}
\usepackage[T1]{fontenc}
\usepackage[utf8]{inputenc}
\usepackage{microtype}
\usepackage{inconsolata}
\usepackage{amssymb}

\usepackage{graphicx}
\usepackage{booktabs}
\usepackage{array}
\usepackage{enumitem}
\usepackage{xspace}
\usepackage{float}
\usepackage{xcolor}
\usepackage{xurl}
\usepackage{amsmath}
\usepackage{comment}
\usepackage{multirow}
\usepackage{listings}

\usepackage{pgfplots}
\pgfplotsset{compat=1.18}

\usepackage{tcolorbox}
\tcbuselibrary{breakable}

\newtcolorbox{PromptBlock}{
    colback=gray!5!white,      
    colframe=gray!50!black,    
    boxrule=0.5pt,             
    arc=2pt,                   
    left=6pt, right=6pt,       
    top=6pt, bottom=6pt,       
    fontupper=\small\ttfamily, 
    breakable                  
}

\title{Can Large Language Models Handle Discourse Particles?\\A Case Study of Colloquial Malay}

\author{ Mariah Al Giptiah Binte Yusoff\thanks{Equal contribution. Authors 1 and 2: conceptualisation, methodology, validation, investigation, resources, data curation, writing -- original draft, and visualisation.} \\ Nanyang Technological University \\ \texttt{l230008@e.ntu.edu.sg} 
\And 
Jakin Tan\footnotemark[1] \\ Nanyang Technological University \\ \texttt{jtan620@e.ntu.edu.sg} 
\AND 
Bocheng Chen\thanks{Authors 3 and 4: software, formal analysis, and validation.} \\ University of Mississippi \\ \texttt{bchen5@olemiss.edu} 
\And 
Guangliang Liu\footnotemark[2] \\ IU Indianapolis\\ \texttt{liugua@iu.edu} 
\And 
Xi Chen\thanks{Corresponding author. Author 5: methodology, validation, investigation, and writing -- review and editing.}\\ Nanyang Technological University \\ \texttt{zoexi.chen@ntu.edu.sg} }

\begin{document}
\maketitle

\begin{abstract}
Discourse particles, such as \textit{well} and \textit{kind of}, are crucial components that enable LLMs to "speak" more like humans. They are used to convey emotions, intentions, and interpersonal attitudes.
However, existing studies have not yet built a comprehensive understanding of LLMs’ capabilities in handling discourse particles. Moreover, the limited number of research focuses primarily on high-resource languages such as English, with little attention paid to Southeast Asian languages. 
In this paper, we (1) propose \textsc{MalayPrag}, a benchmark designed to systematically evaluate and analyze LLMs’ capabilities in handling discourse particles in colloquial Malay; (2) introduce five attributes that provide a theoretically grounded, unified framework for interpreting pragmatic functions of discourse particles. 
Applying these two, we prompt ten off-the-shelf LLMs to perform three prediction tasks. 
The experimental results reveal substantial challenges for current LLMs to accurately connect discourse particles and their pragmatic functions in Malay. The provision of the five attributes designed in this study is found to significantly improve the connections,  highlighting the need for structured scaffolding for models' pragmatic competence.
\textit{Our dataset can be accessed \href{https://github.com/irhama6030/MalayPrag/tree/main}{here}}.
\end{abstract}

\section{Introduction}

With the growing demand for more \textit{human-like} communication of Large Language Models (LLMs) 
an increasing volume of LLM research shows interest in discourse particles for their pragmatic functions, such as expressing hesitation, uncertainty, and nuanced sentiments
~\citep{sheffield2025just,sadlierbrown2024context,wang2025zeroshot,rocha2025crossgenre,eindor2022fortunately}. Pragmatic function—a linguistic concept—refers to the communicative role or purpose that a linguistic expression (in our case, discourse particles) serves in a particular context, including textual functions (e.g., discourse organisation), empressive functions (e.g., stance), and interpersonal functions (e.g., showing politeness)~\cite{aijmer_english_2002}.

A discourse particle is a word or a short expression that does not contribute to the literal meaning of an utterance but serves pragmatic functions 
\citep{grzech2021discourse}. 
Typical examples include \textit{well}, \textit{like}, and \textit{kind of} in English. 
They index pauses, turn-taking, uncertainty, mitigation, identities, and other social qualities~\citep{schiffrin1987discourse,fraser1999discourse}.
\begin{figure*}[t]
    \centering
    \includegraphics[width=0.9\textwidth]{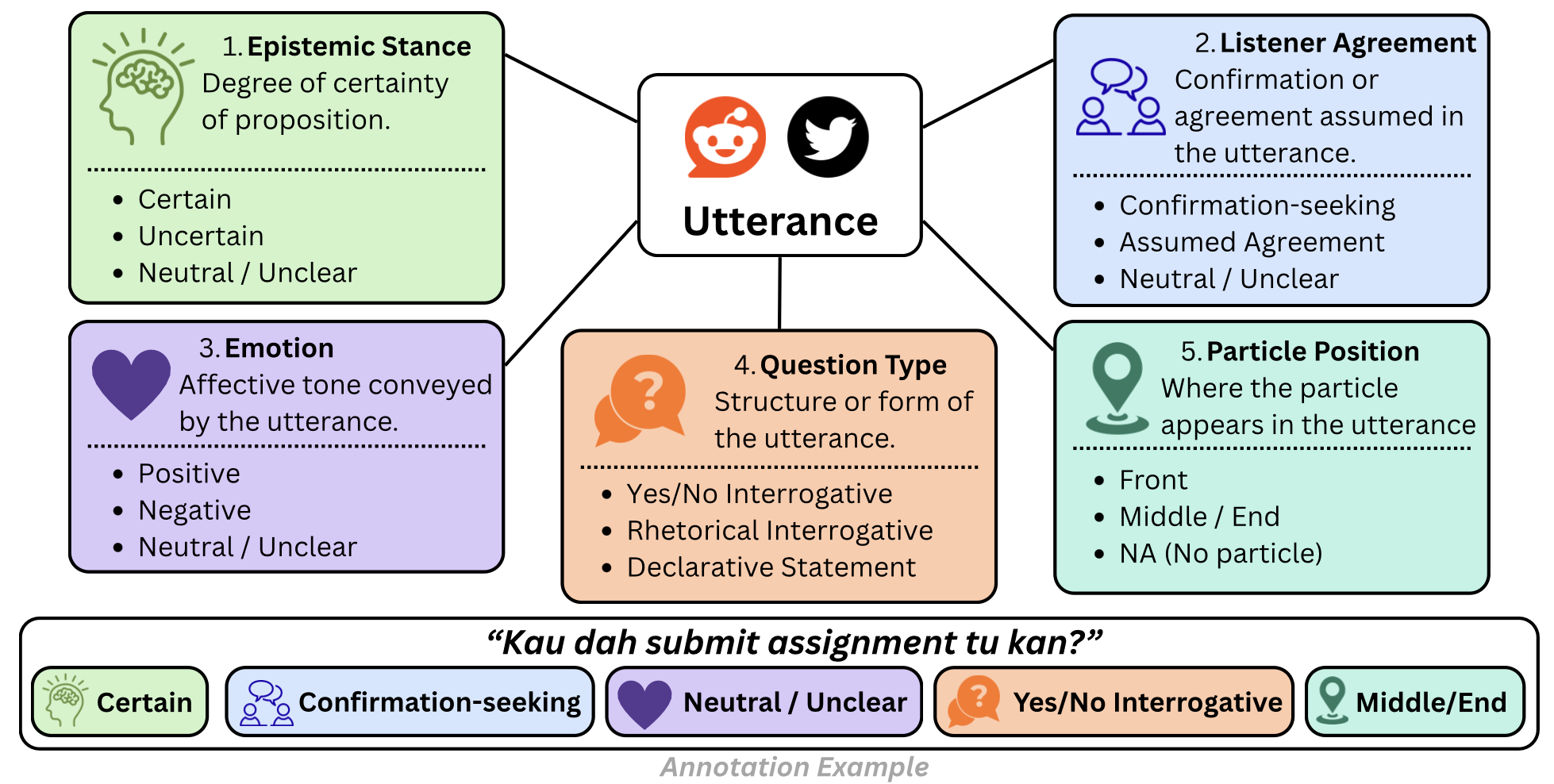}
    \caption{Five-dimensional annotation schema for Malay discourse particle utterances. An utterance is evaluated by a native Malay speaker according to each of the five dimensions and the most appropriate attribute is assigned to the utterance capturing the speaker's intentions and the utterance's syntax. The linguistic theoretical basis for each attribute is available in the Appendix~\ref{tab:flattened-pragmatic-codebook}.}
    \label{fig:details-of-annotations}
\end{figure*}

Understanding LLMs' capability of handling discourse particles and developing methods to improve it are essential for filling the 
well-known pragmatic gaps in LLMs 
~\cite{liu_pragmatic_2025} and enabling them to "speak" more like humans. 
Among existing benchmarks for LLMs' pragmatic performance \citep{sravanthi2024pub,ruis2023large,cong2024manner}, however, very few have touched upon discourse particles, despite the fact that LLMs struggle to capture their pragmatic functions~\cite{sheffield2025just}.
Moreover, these pragmatics-focused studies have primarily examined English and other high-resource languages, leaving low-resource Southeast Asian languages largely unexplored \citep{ma2025pragmatics}. 
To date, no prior work has investigated LLMs' understanding of discourse particles in Malay or its neighboring languages, despite their sophisticated systems of discourse particles.


Given LLMs' struggles with discourse particles, a potential solution is to explore the first-order epistemic, emotional, and linguistic attributes that underlie the varied pragmatic functions, hence translating the abstract notions of functions into discrete, computable variables~\citep{hovy2021importance,choi2023llms}. This study proposes and experimentally examines this solution with LLMs.

In this paper, we not only present a new Colloquial Malay dataset of discourse particles (\textsc{MalayPrag}), but also develop five attributes for evaluating their pragmatic functions in context. The attribute design is solidly grounded in linguistic theories.
Figure~\ref{fig:details-of-annotations} demonstrates the five attributes (\textit{Epistemic Stance}, \textit{Listener Agreement}, \textit{Emotion}, \textit{Question Type}, and \textit{Particle Position}), and their classes for annotation. 

We apply the dataset and attributes to two Malay discourse particles, \textit{kan} and \textit{ke} (commonly spelled as \textit{ka} or \textit{kah}), because they provide a theoretically meaningful contrast in their pragmatic functions (see Section 2 for their details). We conduct three prediction tasks to assess LLMs’ capabilities of handling them: attribute prediction, pragmatic function prediction, and discourse particle prediction, each having a variety of subtasks and measured by model accuracy. We test ten off-the-shelf LLMs.

The experimental results show that \textbf{(1)} LLMs exhibit substantial difficulty in interpreting the pragmatic functions of Malay discourse particles and \textbf{(2)} performance improves when the five attributes are provided (Figure~\ref{fig:details-of-annotations}). Accordingly, the contributions of this study are threefold:
\begin{enumerate}[leftmargin=*, itemsep=0pt, topsep=0pt]
    \item \textit{MalayPrag: A Novel Low-Resource Benchmark.} We introduce a rigorously verified dataset for Colloquial Malay pragmatics, addressing the critical shortage of Southeast Asian representation in LLM evaluation.
    \item \textit{A Theory-Grounded, generalisable framework of five attributes for understanding discourse particle.} We propose five attributes as a unified framework that can be generalised for LLMs to learn pragmatic functions of discourse particles in Southeast Asian languages. \textit{To the best of our knowledge, this is the first attribute framework that enables the unified modeling of pragmatic functions in real-world data.} The attribute-based design also reduces annotator bias and standardizes subjective pragmatic interpretation for computational modeling.
    \item \textit{A fine-grained and comprehensive understanding of LLMs' capability of handling discourse particles.} We provide a detailed comparison of how general-purpose and Southeast Asia-focused LLMs perform on Malay discourse particles in three tasks. Our results show that LLMs' pragmatic competence in low-resource Southeast Asian languages is uneven, attribute-sensitive, and can be improved by explicit pragmatic scaffolding.
\end{enumerate}

\section{Related Work}


In Malay, \textit{kan} and \textit{ke} encode interactional meanings beyond propositional content. \textit{Kan} marks conjoint knowledge or requests listener agreement \citep{tay2016discourse,wouk1998solidarity}, as in ``Dia dah makan, \textbf{kan}?'' (`He already ate, right?'), while \textit{ke} marks interrogativity, uncertainty, confirmation-seeking, or rhetorical challenge, as in ``Awak marah \textbf{ke}?'' (`Are you mad?'). Since these particles vary across epistemic, interpersonal, affective, structural, and positional dimensions, we operationalize their pragmatic functions through five computationally measurable attributes: \textit{Epistemic Stance}, \textit{Listener Agreement}, \textit{Emotion}, \textit{Question Type}, and \textit{Particle Position}. Among them, \textit{Listener Agreement} is designed based on pragmatics studies of common ground and listener orientation \citep{wouk1998solidarity,stalnaker2002common}; epistemic stance and speaker authority motivate the inclusion of \textit{Epistemic Stance} \citep{karkkainen2003epistemic,grzech2021discourse}; and work on affective meaning motivates \textit{Emotion} as a cue to speaker attitude in interaction \citep{caffi1994pragmatics,buechel-hahn-2017-emobank}. Table~\ref{tab:flattened-pragmatic-codebook} summarizes the attribute settings, with full definitions in Appendix~\ref{app:codebook}.

Previous studies have found that LLMs struggle to interpret pragmatic functions of discourse particles ~\citep{sheffield2025just,sadlierbrown2024context,eindor2022fortunately}. For example, \citet{sheffield2025just} demonstrate that LLMs fail to distinguish the overlapping senses of the English particle \textit{just}; moreover, providing surrounding conversational context actively decreases model accuracy rather than aiding disambiguation.

However, it is worth noting that the aforementioned studies primarily used direct, coarse-grained classifications of pragmatic functions, with the hope that LLMs would learn the association between language data and annotated functions automatically. This approach, however, results in severe limitations, including low inter-rater reliability (IRR), data sparseness \citep{defelice2013classification}, and models failing to grasp the underlying reasons why a particle was used. Our approach innovatively bridges the use of discourse particles and their pragmatic functions via the five attributes, by which LLMs' performance improves.


\section{Methodology}

This section details the construction of the \textsc{MalayPrag} dataset. We first outline the data collection and filtering process. Next, we describe the translation of linguistic theory into our annotation schema. Afterwards, we introduce how pragmatic functions were extracted from the data, and 
finally we elaborate the tasks that can be leveraged to understand LLMs' capabilities in handling discourse particles.

\subsection{Data collection}
 Data was sourced from naturally occurring informal Malay utterances on Reddit and Twitter/X containing the particles \textit{kan} and \textit{ke}. To augment the dataset, we synthesized a sample of \textit{kan} sentences to create (1) neutral sentences (by removing \textit{kan}) and (2) substituted utterances (replacing \textit{kan} with \textit{ke}). All synthesized sentences were validated by native speakers to ensure naturalness. The neutral sentences are used to test whether LLMs can distinguish the differences in pragmatic functions with/without the particle.

 With regard to (2) substituted utterances, during dataset construction, we observed that naturally occurring \textit{kan} utterances substantially outnumbered \textit{ke} utterances. Therefore, we generated the substituted utterances to ensure balanced representation across particle use. Crucially, it does not mean that \textit{kan} and \textit{ke} are freely interchangeable; rather, they can occupy identical syntactic positions while communicating different pragmatic intentions (similar to the semantic distinction between 'I can't do it' and 'I won't do it'). This controlled contrast allows us to test a model's ability to discriminate between pragmatic functions independent of syntactic structure.

We used regular expressions to filter Indonesian phrasing that looks like Malay, foreign-language interference, and prepositional uses of the homograph \textit{ke} (`to'), followed by manual inspection for final data  (Appendix E).

\begin{figure}[t]
\centering
\small
\begin{tikzpicture}
\begin{axis}[
    ybar stacked,
    bar width=18pt,
    width=0.48\textwidth,
    height=5cm,
    ymin=0,
    ylabel={Count},
    symbolic x coords={kan, ke, Neutral},
    xtick=data,
    legend style={
        at={(0.5,1.12)},
        anchor=south,
        legend columns=2,
        draw=none
    },
    nodes near coords,
    enlarge x limits=0.25,
]
\addplot coordinates {(kan,63) (ke,60) (Neutral,64)};
\addplot coordinates {(kan,345) (ke,239) (Neutral,366)};

\legend{Gold, Silver}
\end{axis}
\end{tikzpicture}

\caption{
Distribution of Gold and Silver annotation splits across \textit{kan}, \textit{ke}, and neutral baseline utterances. The Gold dataset ($N=187$) is evenly distributed at approximately 33\% per class. Of the total \textit{ke} instances, 113 sentences were synthesized by substituting \textit{kan} with \textit{ke} to address natural imbalance in representation.
}
\label{fig:dataset-distribution}
\end{figure}
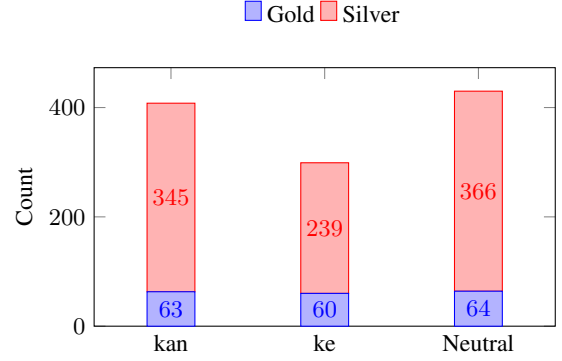

Our research design employed a two-phase pipeline: human annotation of pragmatic attributes followed by computational extraction of pragmatic functions. In the first phase, a total of 1,137 sentences were labeled with our five-attribute schema by native Malay-speaking linguists trained in pragmatic analysis. The resulting dataset was bifurcated into a GOLD and SILVER split to balance the need for an objective baseline against the natural interpretive variability of text-only colloquial Malay. The GOLD dataset ($N=187$) consists of utterances independently annotated by three annotators, with disagreements resolved via majority voting or lead-author adjudication. The SILVER dataset ($N=950$) comprises utterances labeled only by a single trained annotator. In the second phase, both datasets were used to extract the overarching pragmatic functions via K-means clustering. Figure \ref{fig:dataset-distribution} illustrates the data split.

\subsection{Attribute annotation}
Most existing discourse annotation schemes are tailored to English and high-resource languages, making direct porting ineffective for low-resource languages \citep{vargas2025discourse}. Furthermore, applying generic categories without explicit, well-documented definitions leads to subjective bias and high annotator disagreement \citep{crible2015unified}. Therefore, we develop our annotation scheme based on existing linguistic research of Southeast Asian discourse particles and commonly agreed empirical findings across languages (see Section 2 and also Appendix \ref{app:codebook}). 

To be specific, within \textit{Epistemic Stance}, there are three options: Certain, Uncertain, and Neutral. A \textit{Certain} classification denotes that the speaker holds full epistemic authority, presenting the proposition as factual or unquestionable, whereas \textit{Uncertain} reflects speculation, doubt, or an active hypothesis. For \textit{Listener Agreement}, \textit{Assumed Agreement} dictates that the proposition is framed as pre-existing common ground; the speaker expects the listener to simply align or concede, treating the utterance as a rhetorical check rather than a genuine inquiry. Conversely, \textit{Confirmation Seeking} indicates a genuine request for verification where the speaker actively leaves room for the listener to contest, correct, or reject the premise. In contrast, structurally objective variables such as \textit{Particle Position} (\textit{Front, Middle/End, N/A}) and \textit{Question Type} were strictly defined by their syntactic markers and require minimal interpretation.

\subsection{Extracting pragmatic functions from attribute clustering}
As emphasized above, we do not directly annotate pragmatic functions for LLMs to predict, which has been evidenced to be ineffective in previous studies \citep{defelice2013classification}. Instead, we used the five attributes annotated to obtain pragmatic functions. Specifically, we cluster the annotated attributes using K-means on both GOLD ($N=187$) and SILVER ($N=950$) datasets (total $N=1{,}137$). The reason for both datasets to be involved in this process is to take into account both collective understanding of the discourse particles -- that are represented by the agreements in GOLD dataset -- and interpretive variability represented by the SILVER dataset. 

Spatially close clusters of the attributes are expected to represent the same or similar pragmatic functions of a discourse particle, while distanced clusters represent distinctive pragmatic functions. We should emphasise that the clustering results do not reveal the pragmatic functions automatically. Rather, following a K=16 clustering, human linguists conducted a qualitative review of the utterances within the 16 clusters to inductively assign discrete, overarching ``pragmatic function'' labels, thereby establishing a data-driven taxonomy for the subsequent evaluations of LLMs. Table \ref{tab:cluster_mapping} in Appendix \ref{app:particle-functions} shows the distribution of clusters and pragmatic functions.

\paragraph{Association between attributes and pragmatic functions.}
To quantify the relationship between the five annotated attributes and the derived pragmatic-function taxonomy, we conducted Pearson's chi-square tests of independence between pragmatic function and each attribute. We additionally computed Monte Carlo $p$-values to account for potentially sparse contingency-table cells, and Cramér's $V$ as a measure of association strength. All five attributes were significantly associated with pragmatic function (\textit{p} $<$ 2.2e-16 Table~\ref{tab:attribute_function_association}). The strongest associations were observed for Question Type ($V = .893$) and Listener Agreement ($V = .863$), followed by Particle Position ($V = .658$), Epistemic Stance ($V = .626$), and Emotion ($V = .444$). These results confirm the attribute--function associations, with interpersonal and interrogative structure showing particularly strong differentiation across pragmatic functions.

\begin{table}[t]
\centering
\small
\begin{tabular}{lc}
\toprule
\textbf{Attribute} & \textbf{Cramér's $V$} \\
\midrule
Epistemic Stance   & .626 \\
Listener Agreement & .863 \\
Emotion            & .444 \\
Question Type      & .893 \\
Particle Position  & .658 \\
\bottomrule
\end{tabular}
\caption{Association between pragmatic functions and the five annotated attributes. Chi-square tests (Pearson and Monte Carlo) indicated a significant association ($p < 2.2 \times 10^{-16}$). Cramér's $V$ reports the effect size for each attribute.}
\label{tab:attribute_function_association}
\end{table}

 Qualitatively, we also observed the attribute-function association. For example, the pragmatic function ''Information-seeking verification'' is often represented by the attributes of an uncertain epistemic stance, a confirmation-seeking listener agreement, neutral emotion, and an interrogative structure. This specific constellation of attributes indicates that the discourse particle (predominantly \textit{ke}) is functioning as a genuine request for clarification where the speaker actively leaves room for disagreement. Conversely, the ''Negative rhetorical challenge'' function emerges from a sharply contrasting attribute pattern: a certain epistemic stance, assumed listener agreement, a rhetorical question structure, and strong negative affect. In this context, the underlying attributes dictate that the particle functions not as an inquiry, but as a discourse device to criticize, mock, or forcefully evaluate a proposition. Appendix \ref{app:particle-functions} presents representative examples of these derived pragmatic functions.


\begin{table}[t]
\centering
\footnotesize
\renewcommand{\arraystretch}{1.15}
\setlength{\tabcolsep}{2pt}

\begin{tabular}{llcc}
\toprule
\textbf{Prediction} & \textbf{Input} & \textbf{Prompting} & \textbf{+ CoT} \\
\midrule

\shortstack[l]{\textbf{T1} Attribute}
  & Raw data                    & \checkmark & \\
\midrule

\multirow{2}{*}{\shortstack[l]{\textbf{T2} Pragmatic\\Function}}
  & Raw data          & \checkmark & \checkmark \\
  &  + Attributes             & \checkmark & \\
\midrule

\multirow{4}{*}{\shortstack[l]{\textbf{T3} Particle}}
  & Raw data               & \checkmark & \\
  &  + Attributes             & \checkmark & \\
  & + Pragmatic functions    & \checkmark & \checkmark \\
  & + Attr.\ + prag.\ func.  & \checkmark & \\
\bottomrule
\end{tabular}

\caption{Evaluation tasks and conditions. We prompt ten off-the-shelf LLMs to perform three prediction tasks: attribute prediction, pragmatic function prediction, and particle prediction on the \textsc{MalayPrag} benchmark.}
\label{tab:evaluation_tasks}
\end{table}

\subsection{Evaluation Tasks}
To systematically evaluate LLMs' capability to handle the pragmatics of Malay discourse particles, we conduct three evaluation tasks: attribute prediction, pragmatic function prediction, and discourse particle prediction. Table \ref{tab:evaluation_tasks} outlines each task and their conditions. 

\paragraph{Task 1: Attribute Prediction.} 



This task evaluates whether LLMs can predict the five foundational attributes. It comprises two subtasks. In \textbf{(1a) Attribute prediction from raw data}: The model predicts the five attributes relying solely on the provided utterance. In \textbf{(1b) Function-provided attribute prediction}: The model predicts the attributes given the utterance alongside its human-annotated overarching pragmatic function. Comparing these subtasks allows us to explore the bidirectional relationship between attributes and pragmatic functions.

\paragraph{Task 2: Pragmatic Function Prediction.} This task comprises three subtasks. In \textbf{(2a) Function prediction from raw data:} The model predicts predicts the overarching pragmatic function from the utterance alone. In \textbf{(2b) Function prediction via CoT:} The model predicts predicts pragmatic function given utterance and Chain-of-Thought (CoT) \citep{wei2022chain}. In \textbf{(2c) Function prediction via attributes:} The model predicts pragmatic functions given an utterance and its human-annotated attributes. Comparing the three subtasks allows us to confirm the effectiveness of attributes in bridging between discourse particles and their pragmatic functions, especially compared to CoT as a strong baseline.


\paragraph{Task 3: Discourse Particle Prediction.} 

This task evaluates whether LLMs can select the appropriate particle, \textit{kan}, \textit{ke}, or neutral (no particle), for a masked utterance. 
It includes five prompting conditions: \textbf{(3a) Particle prediction from raw data}, using only the masked utterance; \textbf{(3b) Attribute-provided prediction}, adding human-annotated attributes; \textbf{(3c) Function-provided prediction}, adding the target pragmatic function; \textbf{(3d) Particle prediction with attribute + function provided}, adding both; and \textbf{(3e) Particle prediction with CoT + function}, adding pragmatic functions with CoT prompts. Comparing these conditions tests whether explicit pragmatic scaffolding improves particle selection.

\section{Experimental Setting and Results}
This section reports the experimental settings and empirical findings for the tasks above.

\paragraph{Models.} We evaluate ten off-the-shelf LLMs that span varying parameter scales, training paradigms, and regional specialisation. Eight are general-purpose frontier models accessed via their official APIs: GPT-5 and GPT-5.4-mini (OpenAI), Claude Sonnet 4.6 and Claude Haiku 4.5 (Anthropic), Gemini 3.1 Pro and Gemini 3.1 Flash (Google), and DeepSeek-v4-Pro and DeepSeek-v4-Flash (DeepSeek). To examine whether regional training affects pragmatic competence in Malay, we additionally include two open-weight Southeast Asia-focused models from the SEA-LION family~\citep{ng2025sea,koh2025mitigating}: Llama-SEA-LION-70B and Gemma-SEA-LION-27B. 

\paragraph{Evaluation data.} All tasks are conducted on the benchmark \textsc{MalayPrag} ($N=187$), in which every utterance has been independently annotated by three trained native Malay linguists, with disagreements resolved through majority voting or lead-author adjudication. 

\paragraph{Prompting and inference.} Zero-shot, attribute-provided, function-provided, and Chain of Thought ~\cite[CoT]{wei2022chain} are applied, depending on the tasks. 
To minimize sampling variance, we set temperature to $0$ (greedy decoding) for all models and constrain outputs to a single label from the closed set specified by each prompt template. The full prompting templates for every task are provided in Appendix~\ref{app:prompt_templates}.

\paragraph{Evaluation metric.} Following prior work on pragmatic benchmarking~\citep{sravanthi2024pub,sheffield2025just}, we report classification accuracy against the benchmark \textsc{MalayPrag}. For attribute prediction Task, accuracy is computed per attribute and then averaged across the five attributes. For pragmatic function prediction tasks, accuracy is computed over the seven pragmatic-function labels; for particle prediction Tasks, accuracy is computed over the three particle choices (\textit{kan}, \textit{ke}, neutral).  The corresponding random-chance baselines are approximately $33\%$ or $50\%$ for attribute prediction, $14.3\%$ for pragmatic-function prediction, and $33.3\%$ for particle generation.

\begin{table*}[t]
    \centering
    \tiny
    \setlength{\tabcolsep}{2pt}
    \resizebox{\textwidth}{!}{%
    \begin{tabular}{lcccccccccccc}
        \toprule
        \textbf{Model} & \multicolumn{2}{c}{\textbf{Epistemic}} & \multicolumn{2}{c}{\textbf{Agreement}} & \multicolumn{2}{c}{\textbf{Emotion}} & \multicolumn{2}{c}{\textbf{Question}} & \multicolumn{2}{c}{\textbf{Position}} & \multicolumn{2}{c}{\textbf{Overall Avg.}} \\
        \cmidrule(lr){2-3} \cmidrule(lr){4-5} \cmidrule(lr){6-7} \cmidrule(lr){8-9} \cmidrule(lr){10-11} \cmidrule(lr){12-13}
        & \textbf{EN} & \textbf{MS} & \textbf{EN} & \textbf{MS} & \textbf{EN} & \textbf{MS} & \textbf{EN} & \textbf{MS} & \textbf{EN} & \textbf{MS} & \textbf{EN} & \textbf{MS} \\
        \midrule
        GPT-5 & 0.706 & 0.765 & 0.594 & 0.524 & 0.797 & 0.738 & 0.695 & 0.711 & 0.901 & 0.884 & 0.742 & 0.724 \\
        GPT-5.4-mini & 0.583 & 0.449 & 0.631 & 0.519 & 0.797 & 0.770 & 0.604 & 0.626 & 0.752 & 0.777 & 0.700 & 0.628 \\
        Claude Sonnet 4.6 & 0.749 & 0.733 & 0.588 & 0.449 & 0.754 & 0.631 & 0.711 & 0.524 & 0.942 & 0.883 & 0.752 & 0.644 \\
        Claude Haiku 4.5 & 0.749 & 0.636 & 0.578 & 0.449 & 0.701 & 0.684 & 0.679 & 0.679 & 0.860 & 0.771 & 0.722 & 0.644 \\
        Gemini 3.1 Pro & 0.759 & 0.802 & 0.476 & 0.433 & 0.743 & 0.727 & 0.711 & 0.668 & 0.818 & 0.826 & 0.721 & 0.691 \\
        Gemini 3.1 Flash & 0.759 & 0.802 & 0.476 & 0.449 & 0.743 & 0.727 & 0.711 & 0.684 & 0.818 & 0.843 & 0.721 & 0.701 \\
        DeepSeek-v4-Pro & 0.615 & 0.668 & 0.465 & 0.476 & 0.749 & 0.727 & 0.695 & 0.674 & 0.942 & 0.934 & 0.696 & 0.696 \\
        DeepSeek-v4-Flash & 0.706 & 0.722 & 0.476 & 0.438 & 0.754 & 0.754 & 0.679 & 0.690 & 0.942 & 0.884 & 0.713 & 0.698 \\
        \midrule
        Llama-SEA-LION-70B & 0.348 & 0.433 & 0.278 & 0.433 & 0.540 & 0.369 & 0.406 & 0.417 & 0.868 & 0.554 & 0.499 & 0.441 \\
        Gemma-SEA-LION-27B & 0.717 & 0.679 & 0.524 & 0.529 & 0.690 & 0.711 & 0.615 & 0.599 & 0.612 & 0.722 & 0.660 & 0.648 \\
        \midrule
        \textbf{Average} & 0.669 & 0.669 & 0.509 & 0.470 & 0.727 & 0.684 & 0.651 & 0.627 & 0.845 & 0.808 & 0.693 & 0.652 \\
        \bottomrule
    \end{tabular}%
    }
    \caption{Attribute Prediction accuracy for English (EN) and Malay (MS) prompts. Position achieves the highest accuracy while Listener Agreement is the lowest.}
    \label{tab:task-1a-micro-attribute-en-ms}
\end{table*}

\subsection{Task 1: Attribute Prediction}
We use zero-shot English and Malay prompts, separately, to test whether the selected models can accurately predict the five attributes annotated. 
To reiterate, these five attributes underlie our identification of pragmatic functions (see Section 3.3). 

As shown in Table \ref{tab:task-1a-micro-attribute-en-ms}, \textbf{model performance on English (EN) prompts consistently outpaces performance on native Malay (MS) prompts across all models.} The overall average for English prompts was 69.26\%, compared with 65.16\% for Malay prompts. The consistent deficient performance in Malay may be an indication of imbalanced data size and semantic distributions learned during model training. That is, current models may possess Malay vocabulary for processing the task, but the highly technical, meta-linguistic reasoning required to evaluate concepts such as ``epistemic stance'' is overwhelmingly concentrated in English data. 

Among the attributes, Particle Position achieved the highest accuracy (EN: 84.54\%, MS: 80.78\%), confirming that models can execute relatively objective structural and spatial parsing. Conversely, Listener Agreement yielded the poorest performance across the board (EN: 47.0\%, MS: 50.85\%), frequently operating at or below a random chance. Intriguingly, this finding is consistent with human annotators' perceptions of Listener Agreement: This attribute also elicited the highest degree of disagreement among our human annotators during the annotation process. Thus, we argue that \textbf{the lower accuracy in models' prediction of \textit{Listener Agreement} reflects an inherent linguistic ambiguity.} That is, in spoken Malay, assessing common knowledge and listener-oriented assumptions relies heavily on prosody, intonation, and shared conversational history ~\citep{tay2016discourse,wong2004particles}. In the text-only vacuum of social media, where these multi-modal and multi-turn cues are stripped away, modeling perception regarding the listener becomes more challenging for both human linguists and computational models.

Finally, 
we also test whether providing pragmatic functions improves models' prediction of the five attributes (Task 1b). 
As detailed in Appendix \ref{app:task_1b}, this top-down scaffolding yielded consistent performance gains, reflecting the strong statistical dependency between the two linguistic properties.

\subsection{Task 2: Pragmatic Function Prediction}

For the ease of identifying changes in model performance, we first compare pragmatic functions predicted with and without attributes provided. Then, we compare the effectiveness of attributes to that of CoT in function prediction.

\begin{table}[t]
    \centering
    \footnotesize
    \renewcommand{\arraystretch}{1.15}
    \setlength{\tabcolsep}{3pt}
    \begin{tabular}{lccc}
        \toprule
        \textbf{Model} & \shortstack{\textbf{Direct} \\ \textbf{Prompting}}  & \textbf{Attribute} & \textbf{Delta} \\
        \midrule
        GPT-5 & .300 & .529 & .230 \\
        GPT-5.4-mini & .294 & .503 & .209 \\
        Claude Sonnet 4.6 & .257 & .524 & .267 \\
        Claude Haiku 4.5 & .283 & .524 & .241 \\
        Gemini 3.1 Pro & .305 & .519 & .214 \\
        Gemini 3.1 Flash & .337 & .524 & .187 \\
        DeepSeek-v4-Pro & .316 & .588 & .273 \\
        DeepSeek-v4-Flash & .353 & .546 & .193 \\
        \midrule
        Llama-SEA-LION-70B & .123 & .401 & .278 \\
        Gemma-SEA-LION-27B & .230 & .588 & .358 \\
        \midrule
        \textbf{Average} & .280 & .525 & .245 \\
        \bottomrule
    \end{tabular}

    \caption{Pragmatic Function prediction accuracy across models under two conditions: direct prompting (without context) and prompting with attributes. All scores are accuracy; \textbf{Delta is computed as Attribute minus Direct Prompting}. Overall, providing attributes substantially improves pragmatic function prediction across all models.}
    \label{tab:task-1b-1c-macro-function-accuracy}
\end{table}

\begin{table}[t]
    \centering
    \resizebox{\columnwidth}{!}{%
    \begin{tabular}{lccc}
        \toprule
        \textbf{Model} & \textbf{CoT} & \textbf{Attribute} & \textbf{Delta (+/-)}   \\
        \midrule
        GPT-5 & .396 & \textbf{.529} & .133 \\
        GPT-5.4-mini & .316 & \textbf{.503} & .187 \\
        Claude Sonnet 4.6 & .321 & \textbf{.524} & .267 \\
        Claude Haiku 4.5 & .348 & \textbf{.524} & .241 \\
        Gemini 3.1 Pro & .348 & \textbf{.519} & .214 \\
        Gemini 3.1 Flash & .342 & \textbf{.524} & .187 \\
        DeepSeek-v4-Pro & .326 & \textbf{.588} & .272 \\
        DeepSeek-v4-Flash & .364 & \textbf{.546} & .193 \\
        \midrule
        Llama-SEA-LION-70B & .219 & \textbf{.401} & .182 \\
        Gemma-SEA-LION-27B & .262 & \textbf{.588} & .358 \\
        \midrule
        \textbf{Average} & .324 & \textbf{.525} & .201 \\
        \bottomrule
    \end{tabular}%
    }

    \caption{Pragmatic function prediction accuracy across models under two prompting conditions: chain-of-thought prompting (CoT) and attribute-enhanced input (Attributes). Each model assigns one of seven pragmatic function labels to a sentence. \textbf{Delta is computed as Attribute minus CoT}. Overall, explicit attributes outperform chain-of-thought prompting, suggesting that structured pragmatic cues are more useful than elicited reasoning alone.}
    \label{tab:headline-macro-function-ablation}
\end{table}

As Table \ref{tab:task-1b-1c-macro-function-accuracy} displays, \textbf{the provision of attributes significantly improves models' accuracy in predicting pragmatic functions.} Without the attributes, models struggle significantly to predict pragmatic functions, averaging only 27.96\% accuracy. This result is only slightly more accurate than random guessing. However, when scaffolded with the annotated attributes, model performance increased sharply: average accuracy rose to 52.46\%, producing an average diagnostic delta of +24.49\%. 


Similarly, Table \ref{tab:headline-macro-function-ablation} shows that \textbf{the provision of attributes outperforms the incorporation of CoT in pragmatic function prediction.} Applying CoT yields an average accuracy of 32.4\%, which represents only a marginal improvement over the baseline prediction from raw data (27.9\%). The results remains drastically inferior to Attribute-provided Function Prediction performance (52.5\%), with our attribute framework presenting a significant delta of 20.1\%.

Interestingly, while global models such as GPT-5 and Claude Sonnet 4.6 saw reliable gains, Gemma-SEA-LION-27B exhibited the most extreme improvement, with delta over 30\%. With the provision of the attributes, Gemma-SEA-LION-27B even outperformed GPT-5 (52.94\%) and recorded the highest pragmatic function prediction score alongside DeepSeek-v4-Pro.

\subsection{Task 3: Discourse Particle Prediction}
Task 3 contains five sub-tasks. For the ease of comparison, we again divide them into two tables: Table \ref{tab:task-2-generation-accuracy} compares which of attributes or pragmatic functions is more effective in predicting discourse particles, while Table \ref{tab:headline-particle-generation-ablation} compares attributes to CoT.


\begin{table}[t]
    \centering
    \footnotesize
    \renewcommand{\arraystretch}{1.15}
    \setlength{\tabcolsep}{1pt}

    \begin{tabular}{lcccc}
        \toprule
        \textbf{Model} &
        \shortstack{\textbf{Direct} \\ \textbf{\& Prompting}} &
        \textbf{Attr.} &
        \textbf{Func.} &
        \shortstack{\textbf{Attr.} \\ \textbf{\& Func.}} \\
        \midrule
        GPT-5 & .513 & .690 & .674 & .690 \\
        GPT-5.4-mini & .385 & .604 & .497 & .690 \\
        Claude Sonnet 4.6 & .471 & .578 & .546 & .706 \\
        Claude Haiku 4.5 & .471 & .685 & .497 & .711 \\
        Gemini 3.1 Pro & .428 & .631 & .524 & .690 \\
        Gemini 3.1 Flash & .433 & .631 & .519 & .679 \\
        DeepSeek-v4-Pro & .476 & .685 & .562 & .754 \\
        DeepSeek-v4-Flash & .449 & .668 & .588 & .717 \\
        \midrule
        Llama-SEA-LION-70B & .332 & .471 & .417 & .711 \\
        Gemma-SEA-LION-27B & .337 & .503 & .487 & .562 \\
        \midrule
        \textbf{Average} & .430 & .615 & .531 & .691 \\
        \bottomrule
    \end{tabular}

    \caption{Masked-slot particle prediction accuracy (ke, kan, or neutral) under four prompting conditions: direct prompting without context, attribute-only, pragmatic function-only, and combined attribute and pragmatic function context. Overall, the strongest performance comes from combining attribute and pragmatic function information.}
    \label{tab:task-2-generation-accuracy}
\end{table}

As shown in Table \ref{tab:task-2-generation-accuracy}, when predicting from raw data, models average 43.0\% accuracy, which is the lowest across the three tasks. 
Providing pragmatic functions (3c) raises average accuracy to 53.1\%, while providing models with the attributes (3b), instead of pragmatic functions, yields an even higher accuracy, averaging 61.5\%. Providing both attributes and functions (2d) results in the highest average accuracy of 69.1\%. The findings corroborate our argument above that pragmatic functions alone are insufficient for LLMs to interpret discourse particles in low-resource languages. 


\begin{table}[t]
    \centering
    \resizebox{\columnwidth}{!}{%
    \begin{tabular}{lccc}
        \toprule
        \textbf{Model} & \shortstack{\textbf{  CoT} \\ \textbf{\& Function}}  & \shortstack{\textbf{ Attribute} \\ \textbf{\& Function}} & \shortstack{\textbf{Delta} \\ \textbf{(+/-)}} \\
        \midrule
        GPT-5 & .679 & \textbf{.690} & .011 \\
        GPT-5.4-mini & .679 & \textbf{.690} & .011 \\
        Claude Sonnet 4.6 & \textbf{.743} & .706 & -.037 \\
        Claude Haiku 4.5 & .679 & \textbf{.711} & .032 \\
        Gemini 3.1 Pro & .674 & \textbf{.690} & .016 \\
        Gemini 3.1 Flash & .674 & \textbf{.679} & .005 \\
        DeepSeek-v4-Pro & .738 & \textbf{.754} & .016 \\
        DeepSeek-v4-Flash & .706 & \textbf{.717} & .011 \\
        \midrule
        Llama-SEA-LION-70B & .658 & \textbf{.711} & .053 \\
        Gemma-SEA-LION-27B & \textbf{.658} & .562 & -.096 \\
        \midrule
        \textbf{Average} & .689 & \textbf{.691} & .002 \\
        \bottomrule
    \end{tabular}%
    }

    \caption{Masked-slot particle prediction accuracy under two conditions: chain-of-thought with pragmatic function context (CoT \& Function) and joint attribute plus pragmatic function context (Attribute \& Function). Models predict ke, kan, or neutral. Delta is computed as Attribute \& Function minus CoT Function.  Overall, adding attribute context to pragmatic function context yields a small average improvement over CoT with pragmatic function context.}
    \label{tab:headline-particle-generation-ablation}
\end{table}

Interestingly, \textbf{CoT is more effective in predicting discourse particles than in predicting pragmatic functions in Task 2.} As shown in Table \ref{tab:headline-particle-generation-ablation}, CoT + pragmatic function are compared to attributes + pragmatic function. They achieve similar accuracy, although attributes + functions still outperform by a marginal delta. 

This finding aligns with previous studies that find CoT is less effective in pragmatic reasoning tasks, but better at capturing semantic connections \cite{chen-wang-2025-pragmatic, liu_pragmatic_2025, liu2025pragmatic}. In other words, CoT performing strongly in predicting discourse particles may be because the discourse particles fall in the semantic distribution of CoT steps. Pragmatic functions, on the other hand, are usually the "unsaid" effects created by utterances, which CoT can hardly reason. In both Tasks 2 and 3, however, our design of attributes appears to be the strongest scaffolding for LLMs to achieve better prediction accuracy.

\section{Discussion}

\paragraph{The superiority of attribute-based understanding of discourse particles.}
Across different prediction tasks, attributes have consistently been effective in improving model performance on discourse particles. The findings corroborated previous linguistic insights into how they underlie the great variety of pragmatic functions \cite{wouk1998solidarity, stalnaker2002common, karkkainen2003epistemic, caffi1994pragmatics}. Considering that LLMs struggle to learn pragmatic functions directly from language data and the function annotations alone, the current five attributes as a framework have the potential to become the common ``bridge'' that connects LLMs' generation of discourse particles to recognition of their pragmatic functions, especially in low-resource, Southeast Asian languages. 

\paragraph{Regional LLMs' paradox}
Compared to other general-purpose LLMs, the SEA-LION models, which are trained specifically for Southeast Asian languages, are rather inconsistent in performance. Recall that Gemma-SEA-LION-27B improved the most in (2c) pragmatic function prediction with attributes provided. In Task 3, both Llama- and Gemma-based SEA-LION models are much less accurate than other models. We argue that the gap may lie in the pre- and post-training of the SEA-LION models. As \citet{yu-etal-2026-pragmatic} finds, models' sensitivity to pragmatic cues increases consistently with model and data scale, and post-training further consolidates the gains of pragmatic knowledge. Global models receive more training in both stages than SEA-LION models. 
However, SEA-LION models utilize specialized tokenizers for Southeast Asian languages 
\citep{ng2025sea,koh2025mitigating}. These efforts seem to have paid off in the success of the smaller SEA-LION models in leveraging the five attributes and connecting them with pragmatic functions in Malay. In other words, regional LLMs, albeit falling short in model scale and post-training, may be equipped with local cultural and pragmatic knowledge. By supplying explicit pragmatic scaffolding like our five attributes, the local knowledge can effectively transform raw lexical exposure into actionable, culturally accurate reasoning.

\paragraph{Comparing with other evaluation benchmarks.}
When comparing our findings to existing pragmatic evaluations conducted in English, a stark disparity emerges regarding the impact of model input. Recent studies have demonstrated that even smaller LLMs can achieve moderate to high baseline accuracy in English discourse particle prediction tasks \citep{sheffield2025just,ruis2023large}. In contrast, the state-of-the-art frontier models used in the current study still exhibited severe performance degradation in Colloquial Malay.
This difficulty aligns with previous studies which find that models struggle to predict discourse particles in low-resourced languages, for example, in Arabic  \citep{al2026alps} and in Singlish, where models achieved only 15\% to 65.8\% accuracy \citep{jung2025same}. The findings further emphasize the need for theoretically sound and computationally efficient methods, such as our attribute design, to overcome the imbalance in the accessibility of language data.

\section{Conclusion and Future Work}
In this paper, we introduced a new Colloquial Malay dataset, with five attributes designed as its evaluation metrics and for predicting discourse particles. 
Our findings demonstrate that, without the attributes, even the state-of-the-art models fall seriously short in accurately predicting pragmatic functions of discourse particles. Providing the structured, attribute-level grounding drastically improves model performance, outperforming both zero-shot baselines and CoT as well as other conditions. 
While our work only showcased the design on \textit{kan} and \textit{ke} in Colloquial Malay, the Southeast Asian linguistic landscape as a whole is rich with highly polysemous particles. We encourage future work to make further expansion to encompass a wider array of particles and investigate multi-turn conversational contexts, using the attribute-based framework. 
Finally, researchers seeking to use this dataset for fine-tuning should exercise caution regarding the synthesized utterances. 
The manual alternation of the natural distribution of \textit{kan} and \textit{ke} may bias the model training. Future work must carefully account for the synthesized split to avoid distorting underlying statistical patterns learned from natural discourse.

\section*{Limitations}
This study concerns primarily textual data, with the absence of prosodic cues, which may play an important role in annotating the attributes. As much as discourse particles are frequently used in colloquial language, incorporating prosodic features may further enhance the consistency of attribute annotations and, more importantly, enable tests on multimodal LLMs. At the moment, the current study relied on prompting to evaluate LLMs' capability to interpret discourse particles. Whether other methods, such as fine-tuning, may further facilitate LLMs' acquisition of pragmatic functions over attributes and enhance their human-like discourse particle usage is yet to be explored. 

\section*{Acknowledgements}
We would like to acknowledge the funding support from Nanyang Technological University - URECA Undergraduate Research Programme for this research project.


\bibliography{custom}

\appendix

\begin{table*}[t]
    \centering
    \small
    \renewcommand{\arraystretch}{1.1} 
    \begin{tabular}{c|c|c}
        \toprule
        \textbf{Attribute} & \textbf{Classes} & \textbf{Theoretical Basis} \\
        \midrule
        \textbf{Epistemic Stance} & Certain; Uncertain; Neutral & ~\citet{karkkainen2003epistemic,grzech2021discourse}\\
        
        \textbf{Listener Agreement} & Confirmation Seeking; Assumed Agreement; Neutral & ~\citet{wouk1998solidarity,stalnaker2002common}\\
        
        \textbf{Emotion} & Positive; Negative; Neutral & ~\citet{caffi1994pragmatics, buechel-hahn-2017-emobank} \\ 

        \textbf{Question Type} & Declarative; Yes/No Interrogative; Rhetorical & \citet{hoogervorst2018utterance}\\
        
        \textbf{Particle Position} & Front; Middle/End; N/A & \citet{tay2016discourse} \\
        \bottomrule
    \end{tabular}
    \caption{Five attributes for annotating our Malay discourse particle utterances.}
    \label{tab:flattened-pragmatic-codebook}
\end{table*}
\section{Flattened Codebook Definitions}
\label{app:codebook}

This appendix provides the annotation codebook used by human annotators. Each attribute was annotated using a guiding question and a set of discrete tag options.

\subsection{Epistemic Stance}

\noindent\textbf{Guiding question:} How sure does the speaker sound about the information?

\begin{table}[H]
\centering
\small
\begin{tabular}{p{0.28\linewidth}p{0.62\linewidth}}
\toprule
\textbf{Tag} & \textbf{One-sentence rule} \\
\midrule
Certain & Pick this if the speaker treats the statement as already true. \\
Uncertain & Pick this if the speaker sounds unsure, guessing, or checking. \\
Neutral / Unclear & Pick this if you cannot tell whether the speaker is certain or not. \\
\bottomrule
\end{tabular}
\end{table}
\vspace{-0.75em}

\subsection{Listener Agreement}

\noindent\textbf{Guiding question:} What kind of response does the speaker seem to want?

\begin{table}[H]
\centering
\small
\begin{tabular}{p{0.28\linewidth}p{0.62\linewidth}}
\toprule
\textbf{Tag} & \textbf{One-sentence rule} \\
\midrule
Confirmation Seeking & The speaker genuinely expects a direct ``Yes'' or ``No'' answer. \\
Assumed Agreement & Pick this if the speaker expects or leans toward agreement. \\
Neutral / Unclear & The speaker is not looking for any agreement or response from the listener. \\
\bottomrule
\end{tabular}
\end{table}
\vspace{-0.75em}

\subsection{Emotion / Affect}

\noindent\textbf{Guiding question:} What is the underlying ``vibe'' or emotional payload? Choose the single most dominant emotion.

\begin{table}[H]
\centering
\small
\begin{tabular}{p{0.28\linewidth}p{0.62\linewidth}}
\toprule
\textbf{Tag} & \textbf{One-sentence rule} \\
\midrule
Positive & Pick this if the tone feels positive or approving. \\
Negative & Pick this if the tone feels critical, annoyed, or negative. \\
Neutral / Unclear & Pick this if there is little or no emotional signal. \\
\bottomrule
\end{tabular}
\end{table}
\vspace{-0.75em}

\subsection{Question Type}

\noindent\textbf{Guiding question:} Is this a real question, a rhetorical one, or a statement?

\begin{table}[H]
\centering
\small
\begin{tabular}{p{0.28\linewidth}p{0.62\linewidth}}
\toprule
\textbf{Tag} & \textbf{One-sentence rule} \\
\midrule
Yes/No Interrogative & Pick this if the speaker is genuinely asking for an answer. \\
Rhetorical Interrogative & Pick this if it looks like a question but is making a point. \\
Declarative / Statement & Pick this if it is mainly stating something, not asking. \\
Unclear & Pick this if the function is ambiguous. \\
\bottomrule
\end{tabular}
\end{table}
\vspace{-0.75em}

\begin{samepage}
\subsection{Particle Position}

\noindent\textbf{Guiding question:} Where is the particle located?

\vspace{0.25em}
\noindent\begin{minipage}{\linewidth}
\centering
\small
\begin{tabular}{p{0.28\linewidth}p{0.62\linewidth}}
\toprule
\textbf{Tag} & \textbf{One-sentence rule} \\
\midrule
Front & Pick this if the particle appears at the start of the sentence. \\
Middle/End & Pick this if the particle appears anywhere else. \\
N/A & Pick this if no particle is present. \\
\bottomrule
\end{tabular}
\end{minipage}
\end{samepage}

\section{Pragmatic Function Labels}
\label{app:particle-functions}

\subsection{Definitions of pragmatic functions}

This appendix lists the seven pragmatic function labels derived from the clustering analysis.

\begin{description}[leftmargin=!, labelwidth=0.33\linewidth, style=nextline, itemsep=0.25em, topsep=0.25em, parsep=0em]

\item[Assumed-Agreement Rhetorical Stance]
Speaker presents the proposition as already obvious or shared knowledge; the listener is expected to align rather than genuinely answer.

\item[Neutral Declarative]
Plain informational statements with minimal discourse pressure or stance marking.

\item[Information-Seeking Verification]
Genuine request for verification or clarification; the speaker leaves room for disagreement.

\item[Affective Confirmation-Seeking Question]
Speaker seeks confirmation while simultaneously expressing affect, such as surprise, irritation, humor, excitement, or disbelief.

\item[Emphatic / Discourse-Marking]
Particle functions less as a literal confirmation marker and more as a discourse-management or emphasis device.

\item[Null Form Retaining Particle-Like Pragmatic Meaning]
Pragmatic meaning associated with particles remains inferable even after overt particle removal.

\begin{table*}[t]
\centering
\small
\begin{tabular}{@{} >{\raggedright\arraybackslash}p{4.5cm} c p{1.5cm} r >{\raggedright\arraybackslash}p{6cm} @{}}
\toprule
\textbf{Pragmatic Function} & \textbf{Total} & \textbf{Cluster} & \textbf{Count} & \textbf{Representative Example} \\
\midrule
\multirow{3}{=}{Assumed agreement rhetorical stance} & \multirow{3}{*}{339} & Cluster 0 & 306 & \multirow{3}{=}{\textit{``Macam gambar Raya keluarga lah kan. Different categories...''}} \\
 & & Cluster 6 & 30 & \\
 & & Cluster 9 & 3 & \\
\midrule
Neutral declaratives & 302 & Cluster 1 & 302 & \textit{``Micro analyzing, over analyzing bullsh*t betul korang ni support jelah dedua boleh''} \\
\midrule
\multirow{3}{=}{Information-seeking verification} & \multirow{3}{*}{210} & Cluster 4 & 147 & \multirow{3}{=}{\textit{``pergi ke?''}} \\
 & & Cluster 10 & 50 & \\
 & & Cluster 13 & 13 & \\
\midrule
\multirow{3}{=}{Affective confirmation-seeking question} & \multirow{3}{*}{67} & Cluster 2 & 45 & \multirow{3}{=}{\textit{``...memang betul la kita serah lebih 5,000 hektar kepada Indonesia......kan?''}} \\
 & & Cluster 14 & 18 & \\
 & & Cluster 15 & 4 & \\
\midrule
Emphatic / discourse marking & 87 & Cluster 3 & 87 & \textit{``Yang terpilih tu mesti seronok kan, dapat mengubat rindu pada masjid.''} \\
\midrule
\multirow{2}{=}{Null forms retaining pragmatic meaning} & \multirow{2}{*}{82} & Cluster 5 & 75 & \multirow{2}{=}{\textit{``...perubahan pada reka bentuk. Tak sabar nak beli?''}} \\
 & & Cluster 12 & 7 & \\
\midrule
\multirow{3}{=}{Negative rhetorical challenge / evaluation} & \multirow{3}{*}{50} & Cluster 7 & 10 & \multirow{3}{=}{\textit{``Kenapa tak berani persoal chai? Kawan baik ke?''}} \\
 & & Cluster 8 & 17 & \\
 & & Cluster 11 & 23 & \\
\bottomrule
\end{tabular}
\caption{Distribution of 1,137 utterances across the 16 intermediate k-means clusters and their mapping to the 7 pragmatic functions, alongside representative examples.}
\label{tab:cluster_mapping}
\end{table*}

\item[Negative Rhetorical Challenge / Evaluation]
Speaker uses rhetorical questioning to criticize, challenge, mock, or negatively evaluate a proposition rather than genuinely seek information.

\end{description}

\subsection{Attribute clustering for pragmatic function identification} 

(Table 9)

\section{Evaluation Prompt Templates\label{app:prompt_templates}}

\subsection{Benchmark Prompt Examples}

\subsubsection{Task 1a: Attribute Prediction}

\paragraph{Attribute 1: Epistemic Stance}
\begin{PromptBlock}
You are a linguist specializing in colloquial Malay. Your task is to read the Malay sentence below and decide how certain the speaker sounds about the information they are conveying.

Referring to the following three labels and their definitions to make your decision:

Certain: The speaker treats the statement as already true or established. There is no hedging, no doubt, and no checking. The speaker is asserting the information with full confidence.

Uncertain: The speaker sounds unsure, is making a guess, is estimating, or is checking whether something is the case. Words like "agaknya" (I think/probably), "kot" (maybe), or question particles that probe for confirmation are typical signals.

Neutral/NA: The sentence does not carry any detectable certainty signal in either direction. This applies to neutral descriptions, commands, or sentences where certainty is simply not relevant.

Speaker: "\{TEXT\}"

Given the three labels "Certain, Uncertain, Neutral/NA", the most likely, single label of the speaker's utterance is:
\end{PromptBlock}

\paragraph{Attribute 2: Particle Position}
\begin{PromptBlock}
You are a linguist specializing in colloquial Malay. Your task is to read the Malay sentence below and decide where the discourse particle appears in it.

Referring to the following three labels and their definitions to make your decision:

Front: The particle appears at the very start of the sentence, before any other content words.

Middle/End: The particle appears anywhere other than the front -- mid-sentence, before the final word, or at the end.

N/A: No discourse particle is present in the sentence (e.g. the particle slot is shown as "[\_\_\_]" or the sentence simply contains no particle).

Speaker: "{TEXT}"

Given the three labels "Front, Middle/End, N/A", the most likely, single label of the speaker's utterance is:
\end{PromptBlock}

\paragraph{Attribute 3: Listener Agreement}
\begin{PromptBlock}
You are a linguist specializing in colloquial Malay. Your task is to read the Malay sentence below and decide how the speaker is orienting toward the listener in terms of shared knowledge or agreement.

Referring to the following three labels and their definitions to make your decision:

Assumed Agreement: The speaker treats the information as already shared or obvious to the listener. The sentence is presented as common ground -- the underlying tone is "you already know this" or "of course this is true". No explicit confirmation is being requested.

Confirmation Seeking: The speaker is actively checking whether the listener agrees, knows, or can confirm the information. The sentence invites or requests the listener's validation before the speaker can proceed with confidence.

Neutral/Unclear: The sentence does not show any clear orientation toward listener agreement. This applies to plain statements, commands, or cases where the interpersonal stance toward agreement is ambiguous or absent.

Speaker: "{TEXT}"

Given the three labels "Assumed Agreement, Confirmation Seeking, Neutral/Unclear", the most likely, single label of the speaker's utterance is:
\end{PromptBlock}

\paragraph{Attribute 4: Emotion}
\begin{PromptBlock}
You are a linguist specializing in colloquial Malay. Your task is to read the Malay sentence below and decide the emotional tone of the speaker.

Referring to the following three labels and their definitions to make your decision:

Positive: The speaker expresses happiness, excitement, enthusiasm, satisfaction, humor, affection, relief, or any other clearly positive feeling. This includes light-hearted teasing or playful sarcasm that is warm in tone.

Negative: The speaker expresses frustration, annoyance, disappointment, sadness, anger, bitterness, or any other clearly negative feeling. This includes hostile or bitter sarcasm.

Neutral/Unclear: The sentence carries no detectable emotional charge in either direction, or the emotion is genuinely ambiguous and cannot be reliably classified as positive or negative.

Speaker: "{TEXT}"

Given the three labels "Positive, Negative, Neutral/Unclear", the most likely, single label of the speaker's utterance is:
\end{PromptBlock}

\paragraph{Attribute 5: Question Type}
\begin{PromptBlock}
You are a linguist specializing in colloquial Malay. Your task is to read the Malay sentence below and decide its primary sentence function.

Referring to the following three labels and their definitions to make your decision:

Declarative/Statement: The sentence makes an assertion or conveys information. It describes a situation, states a fact, or expresses a view. It is not structured as a question, even if it ends with a particle.

Rhetorical Interrogative: The sentence is phrased as a question but does not expect a genuine answer from the listener. It is used to make a point, express emotion, or emphasize something -- the speaker already implies the answer through the question itself.

Yes/No Interrogative: The sentence is a genuine question that invites the listener to confirm or deny something. The speaker does not already know the answer and is seeking a real yes-or-no response.

Speaker: "{TEXT}"

Given the three labels "Declarative/Statement, Rhetorical Interrogative, Yes/No Interrogative", the most likely, single label of the speaker's utterance is:
\end{PromptBlock}

\subsubsection{Task 1b: Attribute-unprovided Function Prediction}

\paragraph{System message}
\begin{PromptBlock}
You are a linguist specializing in colloquial Malay discourse pragmatics. You must output exactly one label from the provided list and nothing else.
\end{PromptBlock}

\paragraph{User message}
\begin{PromptBlock}
You are a linguist specializing in colloquial Malay discourse pragmatics. Your task is to read the Malay sentence below and identify the primary communicative role the utterance plays in interaction, beyond its literal propositional content.

Referring to the following seven labels and their definitions to make your decision:

Assumed-Agreement Rhetorical Stance: Speaker presents proposition as already obvious/shared knowledge; listener is expected to align rather than genuinely answer.

Neutral Declarative: Plain informational statements with minimal discourse pressure or stance marking.

Information-Seeking Verification: Genuine request for verification or clarification; speaker leaves room for disagreement.

Affective Confirmation-Seeking Question: Speaker seeks confirmation while simultaneously expressing affect (surprise, irritation, humor, excitement, disbelief, etc.)

Emphatic / Discourse-Marking: Particle functions less as a literal confirmation marker and more as a discourse-management or emphasis device.

Null Form Retaining Particle-Like Pragmatic Meaning: Pragmatic meaning associated with particles remains inferable even after overt particle removal.

Negative Rhetorical Challenge / Evaluation: Speaker uses rhetorical questioning to criticise, challenge, mock, or negatively evaluate a proposition rather than genuinely seek information.

Speaker: "Menggatal/Tuaran no electric since 8:50pm. Ada ke notis dari tadi sy check fb sesb tda notis. Tu karen kan cukup2 utk buka 1 lampu ja"

Considering what the speaker is communicatively doing with this utterance, their stance, their orientation toward the listener, and the function the sentence serves in interaction, which of the seven labels best captures its primary discourse role?

The most likely, single label is:
\end{PromptBlock}

\subsubsection{Task 1c: Attribute-provided Function Prediction}

\paragraph{System message}
\begin{PromptBlock}
You are a linguist specializing in colloquial Malay discourse pragmatics. You must output exactly one label from the provided list and nothing else.
\end{PromptBlock}

\paragraph{User message}
\begin{PromptBlock}
You are a linguist specializing in colloquial Malay discourse pragmatics. Your task is to read the Malay sentence below and identify the primary communicative role the utterance plays in interaction, beyond its literal propositional content.

You are provided with the following human-annotated attribute labels for this sentence as additional context:

Certain: The speaker treats the statement as already true or established -- no hedging, no doubt, full confidence.

Middle/End: The particle appears anywhere other than the front -- mid-sentence or at the end.

Neutral/Unclear: No clear orientation toward listener agreement; a plain statement or ambiguous stance.

Neutral/Unclear: No detectable emotional charge, or the emotion is genuinely ambiguous.

Rhetorical Interrogative: Phrased as a question but does not expect a genuine answer; used to make a point or emphasize something.

Use these attributes to inform your decision, but base your final label on the overall communicative function of the utterance.

Referring to the following seven labels and their definitions to make your decision:

Assumed-Agreement Rhetorical Stance: Speaker presents proposition as already obvious/shared knowledge; listener is expected to align rather than genuinely answer.

Neutral Declarative: Plain informational statements with minimal discourse pressure or stance marking.

Information-Seeking Verification: Genuine request for verification or clarification; speaker leaves room for disagreement.

Affective Confirmation-Seeking Question: Speaker seeks confirmation while simultaneously expressing affect (surprise, irritation, humor, excitement, disbelief, etc.)

Emphatic / Discourse-Marking: Particle functions less as a literal confirmation marker and more as a discourse-management or emphasis device.

Null Form Retaining Particle-Like Pragmatic Meaning: Pragmatic meaning associated with particles remains inferable even after overt particle removal.

Negative Rhetorical Challenge / Evaluation: Speaker uses rhetorical questioning to criticise, challenge, mock, or negatively evaluate a proposition rather than genuinely seek information.

Speaker: "Menggatal/Tuaran no electric since 8:50pm. Ada ke notis dari tadi sy check fb sesb tda notis. Tu karen kan cukup2 utk buka 1 lampu ja"

Considering what the speaker is communicatively doing with this utterance, their stance, their orientation toward the listener, and the function the sentence serves in interaction, which of the seven labels best captures its primary discourse role?

The most likely, single label is:
\end{PromptBlock}

\subsubsection{Task 2a: Unprovided Particle Generation}
\begin{PromptBlock}
You are a linguist specializing in colloquial Malay discourse particles. A discourse particle has been removed from the Malay sentence below and replaced with [\_\_\_]. Your task is to predict which particle, either "ke" or "kan" or "neutral", belongs in that slot.

Particle meanings:

ke  : signals genuine uncertainty or invites the listener to confirm something the speaker is unsure about.

kan : signals assumed shared knowledge and seeks light confirmation of something the speaker already believes ("right?").

neutral : indicates no particle.

Speaker: "{TEXT}"

Given the three candidate particles "kan" and "ke" and "neutral", the single most likely particle to fill [\_\_\_] is:
\end{PromptBlock}

\subsubsection{Task 2b: Attribute-provided Particle Generation}

\paragraph{System message}
\begin{PromptBlock}
You are a linguist specializing in colloquial Malay discourse particles. You must output exactly one word -- either "ke" or "kan" or "neutral" -- and nothing else.
\end{PromptBlock}

\paragraph{User message}
Example attributes: Epistemic Stance = Certain, Particle Position = Middle/End, Listener Agreement = Assumed Agreement, Emotion = Negative, Question Type = Rhetorical Interrogative.

\begin{PromptBlock}
You are given a Malay sentence in which one discourse particle has been replaced with [\_\_\_].

Your task is to predict which particle, either "ke," "kan," or "neutral," belongs in the [\_\_\_] slot, such that the discourse-context attributes for this sentence are

The speaker treats the statement as already true or established -- no hedging, no doubt, full confidence.

The particle appears anywhere other than the front -- mid-sentence or at the end.

The speaker treats the information as shared or obvious -- the underlying tone is 'you already know this'.

The speaker expresses frustration, annoyance, disappointment, sadness, anger, or bitterness.

Phrased as a question but does not expect a genuine answer; used to make a point or emphasise something.

Particle meanings:

ke  : signals genuine uncertainty or invites the listener to confirm something the speaker is unsure about.

kan : signals assumed shared knowledge and seeks light confirmation of something the speaker already believes ("right?").

neutral : indicates no particle.

Speaker: "{TEXT}"

Using the sentence context and the attributes above, which single particle -- "ke" or "kan" or "neutral" -- best fills [\_\_\_]?

Return exactly one word from this set and nothing else: ke, kan, neutral
\end{PromptBlock}

\subsubsection{Task 2c: Function-provided Particle Generation}

\paragraph{System message}
\begin{PromptBlock}
You are a linguist specializing in colloquial Malay discourse particles. You must output exactly one word -- either "ke" or "kan" or "neutral" -- and nothing else.
\end{PromptBlock}

\paragraph{User message}
\begin{PromptBlock}
You are given a Malay sentence in which one discourse particle has been replaced with [\_\_\_].
Your task is to predict which particle -- "ke", "kan", or "neutral" -- belongs in the [\_\_\_] slot, such that the sentence is consistent with the primary communicative role the utterance plays in interaction, beyond its literal propositional content:
  Assumed-Agreement Rhetorical Stance: Speaker presents proposition as already obvious/shared knowledge; listener is expected to align rather than genuinely answer.

Particle meanings:

ke: signals genuine uncertainty or invites the listener to confirm something the speaker is unsure about.

kan: signals assumed shared knowledge and seeks light confirmation of something the speaker already believes ("right?").

neutral: indicates no particle.

Speaker: "Menggatal/Tuaran no electric since 8:50pm. Ada [\_\_\_] notis dari tadi sy check fb sesb tda notis. Tu karen kan cukup2 utk buka 1 lampu ja"

Using the sentence context and the macro-function above, which single particle -- "ke" or "kan" or "neutral" -- best fills [\_\_\_]?

Return exactly one word from this set and nothing else: ke, kan, neutral
\end{PromptBlock}

\subsubsection{Task 2d: Attribute \& Function-provided Particle Generation}

\paragraph{System message}
\begin{PromptBlock}
You are a linguist specialising in colloquial Malay discourse particles. You must output exactly one word --- either "ke", "kan", or "neutral" --- and nothing else.
\end{PromptBlock}

\paragraph{User prompt}
\begin{PromptBlock}
You are given a Malay sentence in which one discourse particle has been replaced with [\_\_\_].

Discourse-context attributes for this sentence:
- Epistemic Stance: \{ES\_VAL\} --- \{ES\_DESC\}
- Particle Position: \{PP\_VAL\} --- \{PP\_DESC\}
- Listener Agreement: \{LA\_VAL\} --- \{LA\_DESC\}
- Emotion: \{EM\_VAL\} --- \{EM\_DESC\}
- Question Type: \{QT\_VAL\} --- \{QT\_DESC\}

Macro-pragmatic function of this utterance:
\{MACRO\_FUNCTION\}: \{MACRO\_FUNCTION\_DEFINITION\}

Particle meanings:
  ke      : signals genuine uncertainty or invites the listener to confirm something the speaker is unsure about.
  kan     : signals assumed shared knowledge and seeks light confirmation of something the speaker already believes ("right?").
  neutral : indicates no particle.

Speaker:
  "\{TEXT\_MASKED\}"

Using both the attribute breakdown and the macro-function above, which single particle --- "ke", "kan", or "neutral" --- best fills [\_\_\_]?

Return exactly one word from this set and nothing else: ke, kan, neutral
\end{PromptBlock}

\subsubsection{Task 3a: CoT Attribute-unprovided Function Prediction}

\paragraph{System message}
\begin{PromptBlock}
You are a linguist specialising in colloquial Malay discourse pragmatics. Reason step-by-step about the speaker's communicative intent, then output exactly one macro-function label from the provided list.
\end{PromptBlock}

\paragraph{User prompt}
\begin{PromptBlock}
You are given a Malay sentence. Your task is to classify its primary macro-pragmatic function --- the communicative role the utterance plays in interaction, beyond its literal propositional content.

Referring to the following seven labels and their definitions:

Assumed-Agreement Rhetorical Stance: Speaker presents proposition as already obvious/shared knowledge; listener is expected to align rather than genuinely answer.

Neutral Declarative: Plain informational statements with minimal discourse pressure or stance marking.

Information-Seeking Verification: Genuine request for verification or clarification; speaker leaves room for disagreement.

Affective Confirmation-Seeking Question: Speaker seeks confirmation while simultaneously expressing affect (surprise, irritation, humour, excitement, disbelief, etc.).

Emphatic / Discourse-Marking: Particle functions less as a literal confirmation marker and more as a discourse-management or emphasis device.

Null Form Retaining Particle-Like Pragmatic Meaning: Pragmatic meaning associated with particles remains inferable even after overt particle removal.

Negative Rhetorical Challenge / Evaluation: Speaker uses rhetorical questioning to criticise, challenge, mock, or negatively evaluate a proposition rather than genuinely seek information.

Speaker: "\{TEXT\}"

Think step-by-step:
  1. What is the speaker doing in this utterance (asserting, asking, hedging, emphasising, challenging)?
  2. What is the speaker's stance toward the listener (shared knowledge, seeking confirmation, neutral information, etc.)?
  3. Which of the seven labels best captures this combination?

Format your response EXACTLY as follows, with no extra text before or after:

Reasoning: <step-by-step analysis>
Final answer: <one label from the list above>
\end{PromptBlock}

\subsubsection{Task 3b: CoT Function-Constrained Particle Generation}

\paragraph{System message}
\begin{PromptBlock}
You are a linguist specialising in colloquial Malay discourse particles. Reason step-by-step, then output exactly one particle --- "ke", "kan", or "neutral".
\end{PromptBlock}

\paragraph{User prompt}
\begin{PromptBlock}
You are given a Malay sentence in which one discourse particle has been replaced with [\_\_\_].

The macro-pragmatic function of this utterance is:
  \{MACRO\_FUNCTION\}: \{MACRO\_FUNCTION\_DEFINITION\}

Particle meanings:
  ke      : signals genuine uncertainty or invites the listener to confirm something the speaker is unsure about.
  kan     : signals assumed shared knowledge and seeks light confirmation of something the speaker already believes ("right?").
  neutral : indicates no particle.

Speaker:
  "\{TEXT\_MASKED\}"

Think step-by-step:
  1. What is the speaker communicatively doing, given the macro-function above?
  2. Does the masked slot call for genuine uncertainty (ke), assumed shared knowledge (kan), or no particle (neutral)?
  3. Which choice best fits both the sentence flow and the macro-function?

Format your response EXACTLY as follows, with no extra text before or after:

Reasoning: <step-by-step analysis>
Final answer: <ke | kan | neutral>
\end{PromptBlock}

\subsubsection{Task 3c: Attribute- and Function-Constrained Particle Generation}

\paragraph{System message}
\begin{PromptBlock}
You are a linguist specialising in colloquial Malay discourse particles. You must output exactly one word --- either "ke", "kan", or "neutral" --- and nothing else.
\end{PromptBlock}

\paragraph{User prompt}
\begin{PromptBlock}
You are given a Malay sentence in which one discourse particle has been replaced with [\_\_\_].

Discourse-context attributes for this sentence:
- Epistemic Stance: \{ES\_VAL\} --- \{ES\_DESC\}
- Particle Position: \{PP\_VAL\} --- \{PP\_DESC\}
- Listener Agreement: \{LA\_VAL\} --- \{LA\_DESC\}
- Emotion: \{EM\_VAL\} --- \{EM\_DESC\}
- Question Type: \{QT\_VAL\} --- \{QT\_DESC\}

Macro-pragmatic function of this utterance:
\{MACRO\_FUNCTION\}: \{MACRO\_FUNCTION\_DEFINITION\}

Particle meanings:
  ke      : signals genuine uncertainty or invites the listener to confirm something the speaker is unsure about.
  kan     : signals assumed shared knowledge and seeks light confirmation of something the speaker already believes ("right?").
  neutral : indicates no particle.

Speaker:
  "\{TEXT\_MASKED\}"

Using both the attribute breakdown and the macro-function above, which single particle --- "ke", "kan", or "neutral" --- best fills [\_\_\_]?

Return exactly one word from this set and nothing else: ke, kan, neutral
\end{PromptBlock}

\subsubsection{Task 3b: CoT Function-provided Particle Generation}
\paragraph{System message}
\begin{PromptBlock}
You are a linguist specialising in colloquial Malay discourse particles. Reason step-by-step, then output exactly one particle --- "ke", "kan", or "neutral".
\end{PromptBlock}

\paragraph{User prompt}
\begin{PromptBlock}
You are given a Malay sentence in which one discourse particle has been replaced with [\_\_\_].

The macro-pragmatic function of this utterance is:
  \{MACRO\_FUNCTION\}: \{MACRO\_FUNCTION\_DEFINITION\}

Particle meanings:
  ke      : signals genuine uncertainty or invites the listener to confirm something the speaker is unsure about.
  kan     : signals assumed shared knowledge and seeks light confirmation of something the speaker already believes ("right?").
  neutral : indicates no particle.

Speaker:
  "\{TEXT\_MASKED\}"

Think step-by-step:
  1. What is the speaker communicatively doing, given the macro-function above?
  2. Does the masked slot call for genuine uncertainty (ke), assumed shared knowledge (kan), or no particle (neutral)?
  3. Which choice best fits both the sentence flow and the macro-function?

Format your response EXACTLY as follows, with no extra text before or after:

Reasoning: <step-by-step analysis>
Final answer: <ke | kan | neutral>
\end{PromptBlock}

\subsection{Translated Malay Prompts Examples}
The prompt for Task (1a) Attribute Prediction was translated by the principal author (a native Malay speaker) to test the same models on whether their performance would change with the prompt language.

\subsubsection{Tugas 1a: Ramalan Atribut}

\paragraph{Atribut 1: Pendirian Epistemik}
\begin{PromptBlock}
Anda seorang ahli bahasa yang pakar dalam bahasa Melayu kolokial. Tugas anda adalah untuk membaca ayat Bahasa Melayu di bawah dan menentukan sejauh mana kepastian penutur tentang maklumat yang mereka sampaikan.

Merujuk kepada tiga label berikut dan definisinya untuk membuat keputusan anda:

Pasti: Penutur menganggap pernyataan itu sebagai benar atau sah. Tiada lindung nilai, tiada keraguan, dan tiada semakan. Penutur menegaskan maklumat tersebut dengan penuh keyakinan.

Tidak Pasti: Penutur kedengaran tidak pasti, sedang meneka, sedang menganggarkan atau sedang menyemak sama ada sesuatu itu benar. Perkataan seperti “agaknya”, “kot”, atau partikel soalan yang menyiasat untuk pengesahan adalah isyarat tipikal.

Neutral/Tidak Jelas: Ayat ini tidak membawa sebarang isyarat kepastian yang boleh dikesan sama ada dalam arah. Ini terpakai kepada penerangan, arahan atau ayat neutral di mana kepastian tidak relevan sama sekali.

Penutur: "\{TEKS\}"

Memandangkan tiga label “Pasti, Tidak Pasti, Neutral/Tidak Jelas”, label tunggal yang paling mungkin untuk ujaran penutur ialah:
\end{PromptBlock}

\paragraph{Atribut 2: Kedudukan Partikel}
\begin{PromptBlock}
Anda seorang ahli bahasa yang pakar dalam bahasa Melayu kolokial. Tugas anda adalah untuk membaca ayat Bahasa Melayu di bawah dan menentukan di mana partikel wacana muncul di dalamnya.

Merujuk kepada tiga label berikut dan definisinya untuk membuat keputusan anda:

Depan: Partikel itu muncul pada permulaan ayat, sebelum sebarang perkataan kandungan yang lain.

Tengah/Akhir: Partikel tersebut muncul di mana-mana sahaja selain bahagian hadapan — pertengahan ayat, sebelum perkataan terakhir, atau di hujungnya.

Tiada Partikel: Tiada partikel wacana yang terdapat dalam ayat.

Penutur: "\{TEKS\}"

Memandangkan tiga label “Depan, Tengah/Akhir, Tiada Partikel”, label tunggal yang paling mungkin untuk ujaran penutur ialah:
\end{PromptBlock}

\paragraph{Atribut 3: Persetujuan Pendengar}
\begin{PromptBlock}
Anda seorang ahli bahasa yang pakar dalam bahasa Melayu kolokial. Tugas anda adalah untuk membaca ayat Bahasa Melayu di bawah dan memutuskan bagaimana penutur itu memberi orientasi kepada pendengar dari segi pengetahuan atau persetujuan bersama.

Merujuk kepada tiga label berikut dan definisinya untuk membuat keputusan anda:

Perjanjian Diandaikan: Penutur menganggap maklumat tersebut sebagai telah dikongsi atau jelas kepada pendengar. Ayat tersebut dibentangkan sebagai titik persamaan — nada yang mendasarinya ialah “anda sudah tahu ini” atau “sudah tentu ini benar”. Tiada pengesahan eksplisit diminta.

Mencari Pengesahan: Penutur sedang giat menyemak sama ada pendengar bersetuju, tahu atau boleh mengesahkan maklumat tersebut. Ayat tersebut menjemput atau meminta pengesahan pendengar sebelum penutur boleh meneruskan dengan yakin.

Neutral/Tidak Jelas: Ayat ini tidak menunjukkan sebarang orientasi yang jelas ke arah persetujuan pendengar. Ini terpakai kepada pernyataan, arahan atau kes yang jelas di mana pendirian interpersonal terhadap persetujuan adalah samar-samar atau tiada.

Penutur: "\{TEKS\}"

Memandangkan tiga label “Perjanjian Diandaikan, Mencari Pengesahan, Neutral/Tidak Jelas”, label tunggal yang paling mungkin untuk ujaran penutur ialah:
\end{PromptBlock}

\paragraph{Atribut 4: Emosi / Kesan}
\begin{PromptBlock}
Anda seorang ahli bahasa yang pakar dalam bahasa Melayu kolokial. Tugas anda adalah untuk membaca ayat Bahasa Melayu di bawah dan menentukan nada emosi penuturnya.

Merujuk kepada tiga label berikut dan definisinya untuk membuat keputusan anda:

Positif: Penutur meluahkan kegembiraan, keterujaan, semangat, kepuasan, humor, kasih sayang, kelegaan, atau sebarang perasaan positif yang jelas. Ini termasuk usikan ringan atau sindiran suka bermain yang bernada hangat.

Negatif: Penutur meluahkan rasa kecewa, kegusaran, kekecewaan, kesedihan, kemarahan, kepahitan atau sebarang perasaan negatif yang lain. Ini termasuk sindiran bermusuhan atau pahit.

Neutral/Tidak Jelas: Ayat tersebut tidak membawa sebarang cas emosi yang boleh dikesan dalam kedua-dua arah, atau emosi tersebut benar-benar samar-samar dan tidak boleh diklasifikasikan dengan andal sebagai positif atau negatif.

Penutur: "\{TEKS\}"

Memandangkan tiga label “Positif, Negatif, Neutral/Tidak Jelas”, label tunggal yang paling mungkin untuk ujaran penutur ialah:
\end{PromptBlock}

\paragraph{Atribut 5: Jenis Soalan}
\begin{PromptBlock}
Anda seorang ahli bahasa yang pakar dalam bahasa Melayu kolokial. Tugas anda adalah untuk membaca ayat Bahasa Melayu di bawah dan menentukan fungsi ayat utamanya.

Merujuk kepada tiga label berikut dan definisinya untuk membuat keputusan anda:

Deklaratif/Pernyataan: Ayat ini membuat penegasan atau menyampaikan maklumat. Ia menerangkan situasi, menyatakan fakta atau menyatakan pandangan. Ia tidak berstruktur sebagai soalan, walaupun ia berakhir dengan partikel.

Tanya Jawab Retorik: Ayat ini diungkapkan sebagai soalan tetapi tidak mengharapkan jawapan yang tulen daripada pendengar. Ia digunakan untuk mengemukakan sesuatu perkara, meluahkan emosi atau menekankan sesuatu — penutur sudah menyiratkan jawapan tersebut melalui soalan itu sendiri.

Tanya Jawab Ya/Tidak: Ayat ini merupakan soalan tulen yang mengajak pendengar untuk mengesahkan atau menafikan sesuatu. Penutur belum mengetahui jawapannya dan sedang mencari jawapan ya atau tidak yang sebenar.

Penutur: "\{TEKS\}"

Memandangkan tiga label “Deklaratif/Pernyataan, Tanya Jawab Retorik, Tanya Jawab Ya/Tidak”, label tunggal yang paling mungkin untuk ujaran penutur ialah:
\end{PromptBlock}


\section{Function-Provided Attribute Prediction\label{app:task_1b}}

\begin{table*}[t]
\centering
\small
\begin{tabular}{@{} l c c c c c c @{}}
\toprule
\textbf{Model} & \textbf{Epistemic} & \textbf{Agreement} & \textbf{Emotion} & \textbf{Question} & \textbf{Position} & \textbf{Overall} \\
\midrule
GPT-5 & 0.845 & 0.877 & 0.802 & 0.866 & 0.950 & \textbf{0.868} \\
 & (+0.139) & (+0.283) & (+0.005) & (+0.171) & (+0.049) & \textbf{(+0.130)} \\ \addlinespace

GPT-5.4-mini & 0.765 & 0.813 & 0.775 & 0.834 & 0.934 & \textbf{0.824} \\
 & (+0.182) & (+0.182) & ($-$0.022) & (+0.230) & (+0.182) & \textbf{(+0.151)} \\ \addlinespace

Claude Sonnet 4.6 & 0.882 & 0.840 & 0.786 & 0.807 & 0.975 & \textbf{0.858} \\
 & (+0.133) & (+0.252) & (+0.032) & (+0.096) & (+0.033) & \textbf{(+0.109)} \\ \addlinespace

Claude Haiku 4.5 & 0.807 & 0.658 & 0.738 & 0.813 & 0.926 & \textbf{0.788} \\
 & (+0.058) & (+0.080) & (+0.037) & (+0.134) & (+0.066) & \textbf{(+0.075)} \\ \addlinespace

Gemini 3.1 Pro & 0.909 & 0.706 & 0.781 & 0.834 & 0.975 & \textbf{0.841} \\
 & (+0.150) & (+0.230) & (+0.038) & (+0.123) & (+0.157) & \textbf{(+0.140)} \\ \addlinespace

Gemini 3.1 Flash & 0.914 & 0.690 & 0.775 & 0.829 & 0.975 & \textbf{0.837} \\
 & (+0.155) & (+0.214) & (+0.032) & (+0.118) & (+0.157) & \textbf{(+0.135)} \\ \addlinespace

DeepSeek-v4-Pro & 0.834 & 0.797 & 0.786 & 0.898 & 0.950 & \textbf{0.853} \\
 & (+0.219) & (+0.332) & (+0.037) & (+0.203) & (+0.008) & \textbf{(+0.160)} \\ \addlinespace

DeepSeek-v4-Flash & 0.904 & 0.765 & 0.797 & 0.872 & 0.942 & \textbf{0.856} \\
 & (+0.198) & (+0.289) & (+0.043) & (+0.193) & (+0.000) & \textbf{(+0.144)} \\ \addlinespace
\midrule
Llama-SEA-LION-70B & 0.096 & 0.321 & 0.487 & 0.390 & 0.917 & 0.442 \\
 & ($-$0.252) & (+0.043) & ($-$0.053) & ($-$0.016) & (+0.049) & ($-$0.046) \\ \addlinespace

Gemma-SEA-LION-27B & 0.866 & 0.695 & 0.722 & 0.904 & 0.909 & \textbf{0.819} \\
 & (+0.149) & (+0.171) & (+0.032) & (+0.289) & (+0.297) & \textbf{(+0.188)} \\
\midrule
\textbf{Average} & \textbf{0.782} & \textbf{0.716} & \textbf{0.745} & \textbf{0.805} & \textbf{0.945} & \textbf{0.799} \\
 & \textbf{(+0.113)} & \textbf{(+0.207)} & \textbf{(+0.018)} & \textbf{(+0.154)} & \textbf{(+0.100)} & \textbf{(+0.119)} \\
\bottomrule
\end{tabular}
\caption{Accuracy of predicting five pragmatic attributes when the overarching pragmatic function is provided (Task 1b). The values in parentheses denote the performance difference compared to the raw data baseline (Task 1a).}
\label{tab:reverse_task_1b}
\end{table*}

To address the potential for bidirectional inference, we evaluated whether providing the target pragmatic function improves the models' ability to predict the five lower-level attributes (Task 1b). As shown in Table \ref{tab:reverse_task_1b}, providing the pragmatic function improved attribute prediction accuracy for nearly every model. Nine out of the ten evaluated models saw performance increases, with an average overall gain of +0.137 for English prompts and +0.131 for Malay prompts (indicated by the combined averaged gain). The only exception was Llama-SEA-LION-70B, which experienced a slight degradation. These widespread improvements are unsurprising given the strong statistical association—demonstrated by our Chi-square and Cramér's V analyses (see Section 3.3)—where pragmatic function categorization is highly dependent on these five foundational attributes.

\section{Lexical Filtering Regex}
\label{app:regex}

To identify potentially Indonesian or other non-target language content in the crawled data, we used a manually constructed lexical screening regex. Given the substantial lexical overlap between Malay and Indonesian, the regex was used as a heuristic rather than as a definitive language-identification procedure.

The complete regex is provided below:

\begin{lstlisting}[
basicstyle=\ttfamily\footnotesize,
breaklines=true,
breakatwhitespace=false,
columns=fullflexible,
keepspaces=true
]
\b(aing|aja|ayo|banget|belom|ben|bener|bgt|bilang|bisa|cakep|capek|cewe|cewek|coba|cowo|cowok|de|denger|det|dong|doang|elu|emang|emg|en|enggak|er|ga|gak|gausah|gede|godain|gpp|gua|gue|gueh|gw|gwe|heb|houden|ih|ik|iya|kagak|kalian|kangen|kantor|kemaren|kelmaren|kelmarin|kerjaan|kalo|klo|ko|kok|krna|libur|loe|lu|maen|mau|maar|mensen|mikir|milih|ngapain|ngebadut|ngerebutin|ngelaksanain|ngeband|ngeliat|ngelarang|ngestalk|ngehina|ngerti|ngga|nggak|nih|pake|permen|pulangin|sampe|sebenernya|sih|syuting|temen|temenku|udah|udh|voor|wkwk|yo|zien)\b
\end{lstlisting}

The string \textit{ke} may also occur as the Malay directional 
preposition `to', as part of a larger lexical item, or in other 
non-target expressions. We therefore used a second regex to flag 
likely non-particle occurrences. The text was lowercased prior to 
matching. The filter captured \textit{ke} followed by a set of 
lexical items observed during data inspection, occurrences of 
\textit{ke} embedded within a larger word, and a small set of 
recurring non-target expressions.

\begin{lstlisting}[
basicstyle=\ttfamily\footnotesize,
breaklines=true,
breakatwhitespace=false,
columns=fullflexible,
keepspaces=true
]
\bke (rumah|situ|sana|sini|negara|office|youtube|spotify|iran|reddit|jalan|sabah|mexico|kl|tv1|tv3|komuniti|translation|stress|kepala)\b|[a-z]+ke[a-z]+|\b(je ke|ke je|ke best|ke zelf|ke ni o|wichita, ke|s'ke)\b
\end{lstlisting}

\end{document}